\documentclass[a4paper,fleqn]{cas-dc}

\usepackage{graphicx}
\usepackage{float}
\usepackage{subfig}
\usepackage{xcolor}
\usepackage[numbers,sort]{natbib} 

\def\tsc#1{\csdef{#1}{\textsc{\lowercase{#1}}\xspace}}
\tsc{WGM}
\tsc{QE}

\begin{document}
\let\WriteBookmarks\relax
\def\floatpagepagefraction{1}
\def\textpagefraction{.001}
\let\printorcid\relax 

\shorttitle{LECalib: Line-Based Event Camera Calibration}    

\shortauthors{Liu et al.}

\title[mode = title]{LECalib: Line-Based Event Camera Calibration}  

\nonumnote{}
\author[1]{Zibin Liu}
\ead{Liuzibin@nudt.edu.cn}

\author[1]{Banglei Guan}
\ead{guanbanglei12@nudt.edu.cn} 
\cormark[1] 

\author[1]{Yang Shang}
\ead{shangyang1977@nudt.edu.cn}

\author[2]{Zhenbao Yu}
\ead{zhenbaoyu@whu.edu.cn}

\author[1]{Yifei Bian}
\ead{bianyifei18@nudt.edu.cn}

\author[1]{Qifeng Yu}
\ead{yuqifeng@nudt.edu.cn}

\address[1]{The College of Aerospace Science and Engineering, National University of Defense Technology, Changsha 410073, China}
\address[2]{The GNSS Research Center, Wuhan University, Wuhan 430000, China}

\cortext[1]{Corresponding author} 

\begin{abstract}
	Camera calibration is an essential prerequisite for event-based vision applications. Current event camera calibration methods typically involve using flashing patterns, reconstructing intensity images, and utilizing the features extracted from events. Existing methods are generally time-consuming and require manually placed calibration objects, which cannot meet the needs of rapidly changing scenarios. In this paper, we propose a line-based event camera calibration framework exploiting the geometric lines of commonly-encountered objects in man-made environments, e.g., doors, windows, boxes, etc. Different from previous methods, our method detects lines directly from event streams and leverages an event-line calibration model to generate the initial guess of camera parameters, which is suitable for both planar and non-planar lines. Then, a non-linear optimization is adopted to refine camera parameters. Both simulation and real-world experiments have demonstrated the feasibility and accuracy of our method, with validation performed on monocular and stereo event cameras. The source code is released at~\url{https://github.com/Zibin6/line\_based\_event\_camera\_calib}.
\end{abstract}


%
%


\maketitle

\section{Introduction}

In the past few years, event cameras have attracted a lot of attention from the computer vision and robotics communities~\cite{falanga_davide_dynamic_2020}. There is an enormous commercial interest in exploiting these bio-inspired vision sensors for autonomous drones, mobile robotics, and augmented reality. Event cameras are asynchronous sensors and have unique advantages compared to traditional frame-based cameras: high temporal resolution and low latency, high dynamic range and low power consumption~\cite{gallego_event-based_2022}. Especially in typical scenarios where traditional vision-based methods keep facing challenges, such as navigating fast in complex dynamic environments~\cite{wang2018learning}, challenging illumination conditions~\cite{liu2021relative,DONG2024114088,Dong2020RobustCM}, high-speed sensing~\cite{10121337} and obstacle avoidance~\cite{falanga_davide_dynamic_2020}, event cameras have shown their great potential over other sensing modalities. Before carrying out the aforementioned applications, it is imperative to calibrate the event camera.

\begin{figure}[t] 
	\centerline{\includegraphics[width=7.5cm]{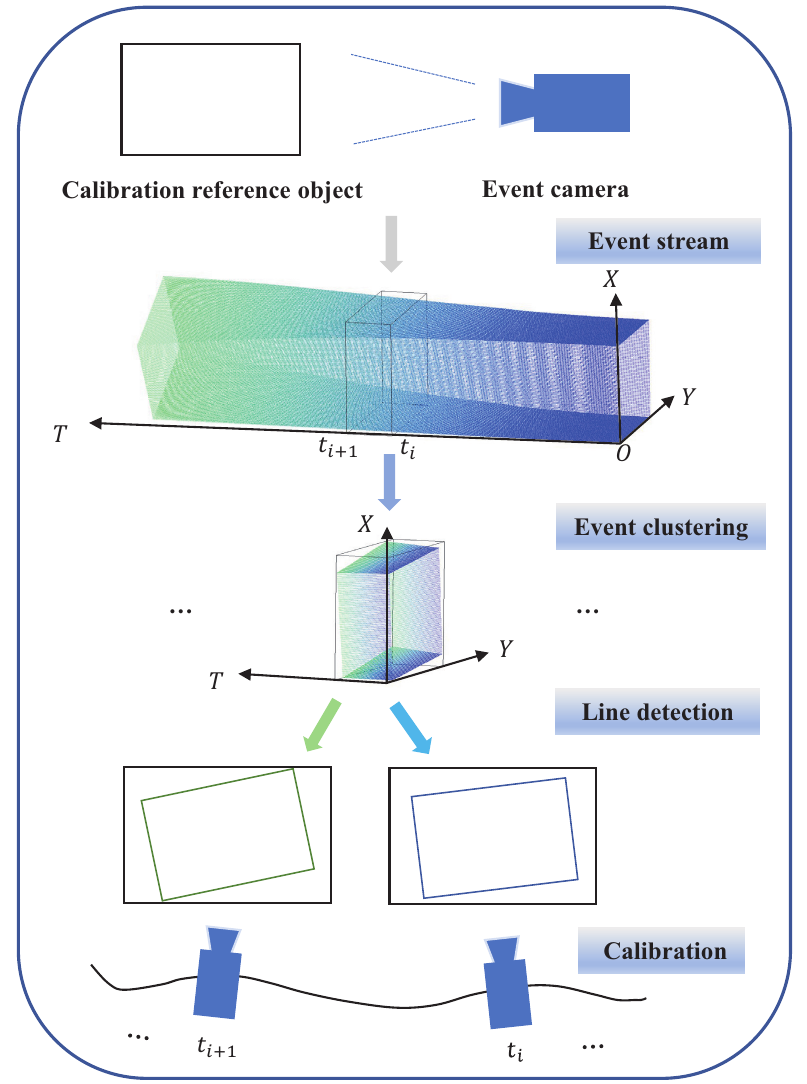}}
	\caption{Overview of line-based event camera calibration framework. Lines are extracted from the events and used for calibration. The linear solution is adopted to provide initial camera parameters, using 2D-3D line correspondences. Then, the non-linear optimization is utilized to refine the parameters by minimizing line-line distance.}
	\label{fig1}
\end{figure}

Camera calibration is an essential and fundamental step in 3D vision applications, as it enables the extraction of metric information from 2D images~\cite{YANG2023113651,ZHANG2018298,Chen2019MirrorassistedPI}. The calibration problem seeks to estimate the camera’s intrinsic parameters (e.g., focal length, principal point, and distortion coefficients) as well as extrinsic parameters (e.g., rotation and translation vectors) from a sufficient number of 2D-to-2D or 3D-to-2D feature correspondences. Similar to conventional cameras, the imaging principles of event cameras can be described by multiple camera models, with the perspective camera model being the most extensively employed among them~\cite{huang_dynamic_2021}. However, traditional calibration methods cannot be directly applied to event cameras since they output asynchronous events rather than traditional images. Event cameras have independent and smart pixels that trigger information asynchronously, encoding the spatial coordinates, temporal information, and polarity. Therefore, we are considering using the raw data from event cameras, combined with their unique characteristics, to conveniently and precisely calibrate the camera.

Current event camera calibration methods, in principle, aim to convert event streams into brightness images and then leverage traditional calibration methods~\cite{muglikar_how_2021}. An alternative is to extract features directly from event streams~\cite{huang_dynamic_2021}. These two methods leverage the features extracted from either images or events for camera calibration, either directly or indirectly. The most classic method for monocular camera calibration is to detect the corners of the checkerboard from different viewpoints~\cite{zhang_flexible_2000}. Nevertheless, an event carries so little information that event-based corners are prone to fail to detect and lose tracking~\cite{liu_edflow_2022}. The precision of corner extraction directly affects the accuracy of camera calibration. Apart from that, these methods often require adjusting the position of the calibration board or capturing images of the board from different angles by moving the camera. These limitations significantly restrict the numerous applications of event cameras, e.g., in high-speed maneuvering scenarios. These issues lead us to wonder if more convenient methods exist for event camera calibration.

To deal with the above problems, we make use of the lines instead of corners for event camera calibration. The main reasons are as follows: Event cameras are sensitive to geometric edges, e.g., lines, because events are triggered mainly by the sharp appearance and scene edges~\cite{peng_globally-optimal_2021}. These contours are able to be modeled and parameterized as a set of lines, which are generally more stable than points, i.e., less likely to be triggered by event noise and outliers~\cite{elised_2016}. On the other hand, man-made environments for vision applications, e.g., robots moving in artificial indoor and outdoor scenarios, contain large amounts of lines. The above reasons make the line an excellent choice for event camera calibration.

In this paper, we fully leverage the potential of the event camera, directly using asynchronous events instead of frame-like event images. Fig. \ref{fig1} provides a visual representation of the process steps involved in our calibration framework. First, we adopt an event-based line detection method that works directly on event streams. 
Second, we linearly solve for the initial values of camera parameters using 2D-3D line correspondences. Subsequently, we utilize the non-linear optimization to further refine the camera parameters.
Finally, we validate the accuracy of our method on both synthetic data and real-world scenes, using monocular and stereo event cameras. To the best of our knowledge, this is the first line-based event camera calibration method. The contributions of LECalib are listed as follows:
\begin{itemize}
	\item We propose a complete line-based event camera calibration framework, including event-based line detection, line-based camera calibration, and non-linear optimization.
	\item Our method works directly on event streams, eliminating the need for grayscale image reconstruction and resulting in a simplified workflow.
	\item Our method removes the requirement for special calibration objects or hardware. Objects commonly found in man-made environments, such as windows, doors, and boxes, can be utilized for calibration purposes. 
\end{itemize}

The remainder of our paper is organized as follows: Section \ref{sec2} reviews related works of event camera calibration and line detection. Section \ref{sec3} presents an overview of the proposed framework, including the details of our camera calibration method. Sections \ref{sec4} and \ref{sec5} respectively validate the proposed method through simulated and real-world experiments. Ultimately, in Section \ref{sec6}, we give the final remarks and simultaneously look forward to future work.

\section{Related work\label{sec2}}
Over the past few years, there has been an increasing number of literature on event-based visual applications. We first review existing literature on event-based line detection and traditional camera calibration methods. Then, we introduce previous works on event camera calibration using different calibration objects. 

Line detection is a fundamental problem in computer vision, which has been extensively researched~\cite{akinlar2011edlines,leavers1993hough,von2012lsd}. One of the most straightforward algorithms for detecting event lines involves aggregating events into an image and subsequently applying traditional line detection algorithms. Among them, hough transform~\cite{conradt2009pencilf} has already been applied for the line detection of events. Event-Based Line Segment Detector (ELiSeD)~\cite{elised_2016}, derived from the LSD method~\cite{von2012lsd}, detects and tracks line segments in an event-by-event fashion. Recently, several line detection works consider events as the 3D point cloud in the space-time volume~\cite{le_gentil_idol_2020-1,lagorce2016hots,reverter_valeiras_event-based_2019}. Then, point cloud-based line detection methods can be applied to extract lines from events. Everding and Conradt~\cite{everding_low-latency_2018} propose a plane fitting method based on Principal Component Analysis (PCA) for line detection and tracking. Xin et al.~\cite{peng_globally-optimal_2021} employ the C++ library Cilantro~\cite{zampogiannis2018cilantro} to implement the line clustering from event streams. The obvious problem with these methods is that events contain a lot of noise and outliers, thereby presenting significant challenges to line extraction. Another issue that remains to be resolved is that the event lines become curved due to lens distortion. As a consequence, there is a pressing need for a robust event-based line detection algorithm.

Camera calibration has been widely investigated in the past, which can provide a bridge between the 2D image and 3D real-world distances. Line features, as the most common structure in the real world, have been utilized for calibration of optical sensors such as monocular cameras~\cite{chuang2021geometry}, stereo cameras~\cite{shi2021dlt}, Time-of-Flight (ToF) cameras~\cite{liu2021three}, etc. Moreover, line features can also intuitively reflect the distortion characteristics of the lens, which can be directly used to solve distortion coefficients. However, as far as we know, there is currently no method that utilizes lines for event camera calibration.

As mentioned in the introduction, the existing works for event camera calibration mainly rely on the usage of blinking calibration targets~\cite{cho2021eomvs,dominguez2019bio,MuegglerEvent6dof}, grayscale image reconstruction~\cite{muglikar_how_2021}, or event-based feature extraction~\cite{huang_dynamic_2021}. Once the frame-like event images are obtained or the 2D features are extracted, traditional methods can be applied for camera calibration. As regards the first category, flashing light~\cite{cho2021eomvs}, LED screens~\cite{dominguez2019bio,reinbacher2017real} or blinking electronic devices~\cite{MuegglerEvent6dof} are probably the most widely employed calibration targets. Events can be triggered and accumulated by the rapid brightness changes, even if the camera and calibration target are relatively static. Manuel et al.~\cite{dominguez2019bio} calibrate the stereo event camera by utilizing a specialized calibration board with 64 LED lights, which are controlled by a PIC microcontroller. Such an approach requires specialized hardware to control the flicker frequency of the LED or electronic devices, which is expensive, complex, and lacks universality. Considering the asynchronous and sparse nature of events, in the second category, Muglikar et al.~\cite{muglikar_how_2021} use image reconstruction of the checkerboard from sparse events for calibration. The former two categories convert events to 2D images that are well suited for traditional calibration algorithms, which may fail to unlock the potential of event cameras. Fundamentally, they do not take full advantage of the continuous-time property of events. Compared with previous approaches, the third category directly extracts features from event streams for calibration. Huang et al.~\cite{huang_dynamic_2021} adopt the circle fitting algorithm to detect circle features from event clusters. Then, a continuous-time back-end optimization is adopted to refine the initial camera parameters. Circles appear to be ellipses under the condition of a large inclined view between the camera and calibration target, which inevitably brings the eccentricity error of circular targets and affects the calibration accuracy.

In this work, we propose a line-based event camera calibration framework that leverages the geometric lines of commonly-encountered objects. Different from the methods mentioned above, we do not require specific calibration boards or flickering electronic devices. Additionally, our method directly operates on event streams without the need to reconstruct grayscale images, thereby simplifying the calibration process.

\section{Line-based event camera calibration\label{sec3}}
We describe the proposed framework for line-based event camera calibration. First, we present the line detection method, which works directly on event streams. Given a set of planar or non-planar line correspondences, we calibrate the event camera to obtain the initial value of camera parameters. Non-linear optimization is adopted to refine the initial parameters. 

\subsection{Event-based line detection}\label{sec3A}
As the asynchronous, bio-inspired visual sensors, event cameras have independent pixels that trigger information in the continuous log brightness signal~\cite{falanga_davide_dynamic_2020}. The camera sends an event when the brightness change at the pixel $\left( x,y \right)$ exceeds a user-defined threshold. Mathematically, each event can be described as $e=\left( x,y,t,s \right)$, where $t$ is the trigger time and $s$ is a binary polarity (brightness increasing (“ON”) or decreasing (“OFF”)). Event streams in a spatio-temporal neighborhood are often treated as a 3D point cloud~\cite{le_gentil_idol_2020-1}, where the first two components are the event’s pixel coordinates. The third coordinate is the event’s trigger time arbitrarily normalized: $\left( x,y,{t}/{c}_{z} \right)\in {{\mathbb{R}}^{3}}$. The time dimension becomes a geometric one, which is a sparse representation. Adjusting the constant ${c}_{z}$ makes the point cloud more compact in the time dimension. On short time scales, lines that leave traces of events in the space-time volume are approximately planar~\cite{everding_low-latency_2018}. Based on this assumption, we adopt an event-based line detection method, which is summarized in the following procedure:

\textbf{Pre-processing:} The first step of our method requires the collection of events over a short time interval for event clustering, which should contain enough events to perform reliable line detection. Then, event clusters are converted into point clouds in the spatio-temporal domain $\left( x,y,{t}/{c}_{z} \right)\in {{\mathbb{R}}^{3}}$. Another benefit of processing events together is that it produces a sufficient signal-to-noise ratio, which can be effectively combined with denoising algorithms~\cite{rusu_towards_2008} to reduce the interference of event noise.

\textbf{Event plane segmentation:} This step aims to split the event cluster into event planes. The K Nearest Neighbour (KNN) method is adopted to find the neighbors of each event, and the PCA algorithm is applied to obtain the normal of the neighboring surface~\cite{lu_fast_2019,lu_pairwise_2016}. The covariance matrix for each event can be obtained by:
\begin{equation} 
	\sum={\frac{1}{s}\sum\nolimits_{i=1}^{s}{\left( {\mathbf{P}_{i}}-\mathbf{\overline{P}}  \right){{\left( {\mathbf{P}_{i}}-\mathbf{\overline{P}} \right)}^{\prime }}}},
	\label{eq1-1}
\end{equation}
where $\sum{{}}$ is the $3 \times 3$ covariance matrix, $\mathbf{P}_{i}$ and $\mathbf{\overline{P}}$ are neighboring points and the mean vector of the KNNs, respectively. The normal of the point can be obtained using Singular Value Decomposition (SVD), which corresponds to the third eigenvector in the matrix of eigenvectors. Then, the region growing and merging strategy aims to extract all 3D planes from the event cluster. The output of this step is a set of event planes along with their corresponding normal vectors ${{n}_{\prod }}$.

\textbf{Plane projection:} Given the extracted event planes, the previous methods directly perform plane fitting~\cite{peng_globally-optimal_2021}. However, these methods are highly sensitive to the noise of the event cluster, which maybe not the best choice. Hence, we conduct plane projection to eliminate redundant information and noise from the events, facilitating the extraction of lines. Another advantage of this step is that it can effectively eliminate the impact of distortion. Camera distortion causes straight lines to curve, leading to the formation of a curved surface in a spatio-temporal neighborhood. Within a short time interval, the edges of a slightly curved plane approximate straight after projection, thereby reducing the interference caused by distortion.

The main steps can be described as follows: Firstly, for each segmented event plane $\mathbf{\prod} $, its belonging events are orthographically projected onto the plane itself, denoted as $\mathbf{{\prod }'} $. We calculate the central point $\mathbf p_c $ of the event plane $\mathbf{\prod} $. Then, we choose a point $\mathbf P_1 $ on the plane $\mathbf{\prod}$, and project it onto the plane $\mathbf{{\prod }'} $, denoting the projection as $\mathbf p_1 $. The x-axis is set to $\overrightarrow{{\mathbf{p}_{c}}{\mathbf{p}_{1}}}$ with the positive direction $v_x$. The y-axis can be solved by:
\begin{equation} 
	{{v}_{y}}={{v}_{x}}\times {{n}_{\prod }}.
	\label{eq1-2}
\end{equation}

For each point $\mathbf{P}_{i}$ on the plane $\mathbf{\prod} $, we project it onto the plane $\mathbf{{\prod }'}$ to obtain $\mathbf{p}_{i}=\left( {{x}_{i}},{{y}_{i}} \right)$ by the following equation:
\begin{equation} 
	\left\{ \begin{matrix}
		\overrightarrow{{{\mathbf{p}}_{c}}{{\mathbf{p}}_{1}}}=\overrightarrow{{{\mathbf{p}}_{c}}{{\mathbf{p}}_{1}}}-\left( \overrightarrow{{{\mathbf{p}}_{c}}{{\mathbf{p}}_{1}}}\cdot {{\mathbf{n}}_{\Pi }} \right){{\mathbf{n}}_{\Pi }},  \\
		{{x}_{i}}=\overrightarrow{{{\mathbf{p}}_{c}}{{\mathbf{p}}_{1}}}\cdot {{v}_{x}},  \\
		{{y}_{i}}=\overrightarrow{{{\mathbf{p}}_{c}}{{\mathbf{p}}_{1}}}\cdot {{v}_{y}}.  \\
	\end{matrix} \right.
	\label{eq1-3}
\end{equation}

After projecting all events onto the plane $\mathbf{{\prod }'} $, we perform line detection and back-project lines into the spatial domain. Generally, most lines in event clusters are extracted.


Following the above processing steps, we obtain two sets of lines almost parallel, corresponding to the object state at the initial and end time of the event cluster, respectively.

\begin{figure}[t] 
	\centerline{\includegraphics[width=8cm]{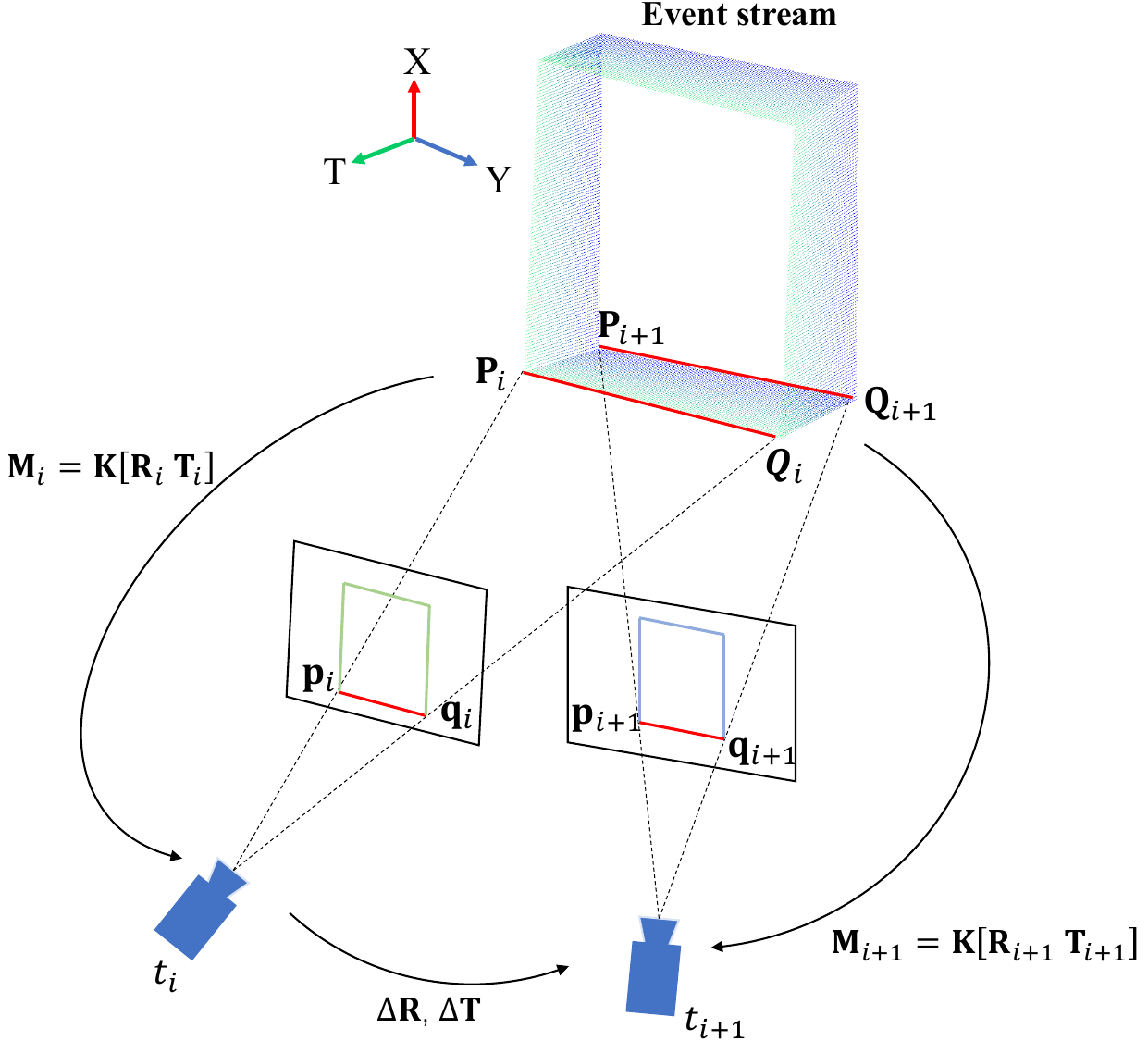}}
	\caption{  Illustration of the 3D line projection and the relative motion. ${\mathbf P_{i}}{\mathbf Q_{i}},{\mathbf P_{i+1}}{\mathbf Q_{i+1}}$ are detected lines at time ${{t}_{i}},{{t}_{i+1}}$, which are parameterized by its two endpoints. The relative motion of the event camera from time ${{t}_{i}}$ to time ${{t}_{i+1}}$ can be described by rotation $\Delta \mathbf R$ and translation $\Delta \mathbf T$.}
	\label{fig2}
\end{figure}

\subsection{Initial calibration\label{sec3B}}
In order to present this matter from a mathematical perspective, let us start with the ideal pinhole camera model, which can be described by a central projection matrix:
\begin{equation} 
	{\mathbf{M}}={\mathbf{K}}[{\mathbf{R}}\text{ }\ {\mathbf{T}}], \label{eq2-1}
\end{equation}
where ${\mathbf{R}}=\left[ {{r}_{ij}} \right]_{i,j=1}^{3}$ is a $3\times 3$ rotation matrix, ${\mathbf{T}}={{\left[ {{t}_{x}},{{t}_{y}},{{t}_{z}} \right]}^{T}}$ is a translation vector, and ${\mathbf{K}}$ is the $3\times 3$ camera intrinsic matrix containing the focal length $\left( {{f}_{x}},{{f}_{\text{y}}} \right)$ and principal point $\left( {{c}_{x}},{{c}_{\text{y}}} \right)$. The problem of event camera calibration resides in estimating the projection matrix $\mathbf{M}$, which encodes the intrinsic and extrinsic parameters. Let an event be $\mathbf{P}_{i}$, and its corresponding 2D projection on the image be $\mathbf{p}_{i}={{\left[ {{u}_{i}}\text{ }{{v}_{i}} \right]}^{T}}$, as shown in Fig. \ref{fig2}. By applying the central projection constraint, we can establish the following relation:
\begin{equation} 
	\lambda \mathbf{p}_{i}={\mathbf{M}_{i}}\mathbf{P}_{i}, \label{eq2-2}
\end{equation}
where $\lambda \in \mathbb{R}$ is an unknown scaling factor. The two endpoints of the line $\mathbf{L}$ are denoted as ($\mathbf{P}_{i},\mathbf{Q}_{i}$) 
, and their projection $\left( \mathbf{p}_{i},{ }\mathbf{q}_{i} \right)$ must lie on the projected line $\mathbf{l}_{i}$.


Pre-normalization is essential to improve the conditioning of the measurement data (\emph{i.e.}, the endpoints of the 3D and 2D lines). We relocate these endpoints in order to align their centroid with the origin. Then, we normalize these endpoints to ensure their average distance to the origin is equal to $\sqrt{2}$ for 2D endpoints and $\sqrt{3}$ for 3D endpoints. After pre-normalization processing, our algorithm is carried out using the normalized data.

For each line correspondence, we have two constraints:
\begin{equation} 
	\left\{ \begin{matrix}
		\mathbf l_{i}^{T}\mathbf{p}_{i}=\mathbf l_{i}^{T}{{\mathbf{M}_{i}}}\mathbf{P}_{i}=0,  \\
		\mathbf l_{i}^{T}\mathbf{q}_{i}=\mathbf l_{i}^{T}{{\mathbf{M}_{i}}}\mathbf{Q}_{i}=0.  \\
	\end{matrix} \right.
	\label{eq2-3}
\end{equation} 

According to the principle of matrix vectorization, Eq. (\ref{eq2-3}) can be transformed into:
\begin{equation} 
	\left\{\begin{array}{l}
		\left(\mathbf P_{i}^{T} \otimes \mathbf l_{i}^{T}\right) \cdot \operatorname{vec}\left({\mathbf{M}_{i}}\right)=0, \\
		\left(\mathbf Q_{i}^{T} \otimes \mathbf l_{i}^{T}\right) \cdot \operatorname{vec}\left({\mathbf{M}_{i}}\right)=0.
	\end{array}\right.
	\label{eq2-4}
\end{equation}

The left-hand side can be replaced by the measurement matrix $\mathbf{A}=\left(\mathbf P_{i}^{T} \otimes \mathbf l_{i}^{T}\right)$, and the vector of unknowns $\mathbf m_{i}=\operatorname{vec}\left(\mathbf{M}_{i}\right)$. In this case, the linear constraints can be obtained by concatenating 2D-3D normalized endpoints:  

\begin{equation}
	\mathbf{A} \mathbf m_{i}=0.
	\label{eq2-5}
\end{equation} 

There are several strategies to solve this linear equation. The most straight-forward method is Direct Linear Transformation (DLT)~\cite{abdel-aziz_direct_2015}, which provides a least square solution of Eq. (\ref{eq2-5}) based on SVD of the measurement matrix $\mathbf{A}$. To improve the computing efficiency for large measurement data, we compute SVD of $\mathbf{A}^{T}\mathbf{A}$, which has the same right singular vectors as $\mathbf{A}$. The smallest eigenvalue is the solution of $\mathbf m_{i}$. Furthermore, it is necessary to apply the inverse transformation to compensate for pre-normalization processing to obtain accurate and reliable results.

For different calibration objects, they can either be planar or non-coplanar. Taking into account different line configurations, the solution of $\mathbf m_{i}$ can be divided into the following two cases:

\textbf{For planar lines:} Without loss of generality, all 3D lines are defined on a common plane. The elements of the projection matrix $\mathbf M_{i}$ are composed as follows:
\begin{equation}
	\mathbf M_{i}=\left[ \begin{matrix}
		{{f}_{x}}{{r}_{1}}+{{c}_{x}}{{r}_{7}} & {{f}_{x}}{{r}_{2}}+{{c}_{x}}{{r}_{8}} & {{f}_{x}}{{t}_{1}}+{{c}_{x}}{{t}_{3}}  \\
		{{f}_{y}}{{r}_{4}}+{{c}_{y}}{{r}_{7}} & {{f}_{y}}{{r}_{5}}+{{c}_{y}}{{r}_{8}} & {{f}_{y}}{{t}_{2}}+{{c}_{y}}{{t}_{3}}  \\
		{{r}_{7}} & {{r}_{8}} & {{t}_{3}}  \\
	\end{matrix} \right].
	\label{eq2-6}
\end{equation}

In this case, the rank of the measurement matrix $\mathbf{A}$ is eight, and each line correspondence provides two independent linear equations~\cite{hartley2003multiple}. For camera calibration, it is often reasonable to assume that the principal point is well approximated by the image center~\cite{nakano2016versatile,penate2013exhaustive}. Hence, if more than four line correspondences are available, Eq. (\ref{eq2-5}) can be solved. Then, the intrinsic and extrinsic parameters of the projection matrix $\mathbf M_{i}$ can be recovered from $\mathbf m_{i}$. In detailed terms, with the known principal point, the focal length $\left( {{f}_{x}},{{f}_{\text{y}}} \right)$ can be uniquely determined based on Eq. (\ref{eq2-6}). Then, according to the orthonormality constraints (unit-norm and mutually orthogonal rows) of the rotation matrix and ${{t}_{z}}>0$, extrinsic parameters can be determined linearly.

\textbf{For non-planar lines:} The elements of the projection matrix $\mathbf M_{i} $ are composed as follows:
\begin{equation}
	\mathbf M_{i} \!=\!\left[ \begin{matrix}
		{{f}_{x}}{{r}_{1}}\!+\!{{c}_{x}}{{r}_{7}} & {{f}_{x}}{{r}_{2}}\!+\!{{c}_{x}}{{r}_{8}} & {{f}_{x}}{{r}_{3}}\!+\!{{c}_{x}}{{r}_{9}} & {{f}_{x}}{{t}_{1}}\!+\!{{c}_{x}}{{t}_{3}}  \\
		{{f}_{y}}{{r}_{4}}\!+\!{{c}_{y}}{{r}_{7}} & {{f}_{y}}{{r}_{5}}\!+\!{{c}_{y}}{{r}_{8}} & {{f}_{y}}{{r}_{6}}\!+\!{{c}_{y}}{{r}_{9}} & {{f}_{y}}{{t}_{2}}\!+\!{{c}_{y}}{{t}_{3}}  \\
		{{r}_{7}} & {{r}_{8}} & {{r}_{9}} & {{t}_{3}}  \\
	\end{matrix} \right].
	\label{eq2-7}
\end{equation}

The intrinsic and extrinsic parameters are encoded into the $3\times 4$ projection matrix $\mathbf M_{i} $. There are twelve entries, and thus at least six line correspondences are able to fully determine the Eq. (\ref{eq2-5}). The decomposition of the projection matrix $\mathbf M_{i} $ is different from the former case. Firstly, ${{\mathbf{M}_{i}}_{\left( 1:3,1:3 \right)}}$ is executed by QR decomposition to obtain the intrinsic matrix $\mathbf{K}$ and the rotation matrix $\mathbf{R}$. Secondly, ensure the focal length is positive. Lastly, the remaining parameter to recover is the translation vector $\mathbf{T}$, which can be decomposed from the fourth column of the matrix projection $\mathbf M_{i}$.

Thus far, we have determined the intrinsic and extrinsic parameters of the event camera. However, traditional cameras usually exhibit significant lens distortion~\cite{zhang_flexible_2000}, especially radial and tangential distortion. The most commonly distortion models include brown model~\cite{duane1971close}, even-order polynomial model~\cite{hartley2003multiple}, and division model~\cite{fitzgibbon2001simultaneous}. Next, we take the brown model as an example: 
\begin{equation}
	\left\{\! \begin{array}{*{35}{l}}
		{{x}_{d}}\!=\!\left( 1\!+\!{{k}_{1}}d_{u}^{2}\!+\!{{k}_{2}}d_{u}^{4}\!+\!{{k}_{5}}d_{u}^{6} \right){{x}_{u}}\!+\!2{{k}_{3}}{{x}_{u}}{{y}_{u}}\!+\!{{k}_{4}}\left( d_{u}^{2}\!+\!2x_{u}^{2} \right),  \\
		{{y}_{d}}\!=\!\left( 1\!+\!{{k}_{1}}d_{u}^{2}\!+\!{{k}_{2}}d_{u}^{4}\!+\!{{k}_{5}}d_{u}^{6} \right){{y}_{u}}\!+\!2{{k}_{4}}{{x}_{u}}{{y}_{u}}\!+\!{{k}_{3}}\left( d_{u}^{2}+2y_{u}^{2} \right).  \\
	\end{array} \right.
	\label{eq2-8}
\end{equation}
Where ${{[{{x}_{u}}, {{y}_{u}}]}^{T}}$ and ${{[{{x}_{d}}, {{y}_{d}}]}^{T}}$ are the corresponding normalized image coordinates of an ideal (undistorted) point and a real (distorted) point, respectively. $k=\left[ {{k}_{1}},{{k}_{2}},{{k}_{3}},{{k}_{4}},{{k}_{5}} \right]$ is the distortion coefficients, and ${{d}_{u}}=\sqrt{x_{u}^{2}+y_{u}^{2}}$ is the Euclidean distance. We project 3D lines onto the image to obtain the projected lines. Next, the distortion parameters are derived linearly by establishing constraints with detected lines in Eq. (\ref{eq2-8}). Besides, the alternative to obtain the distortion coefficients is to set the initial coefficients $k$ to zero, and then use the non-linear optimization, as we will introduce in Section \ref{sec3C}. Without loss of generality, our method can be extended to other distortion models.

\subsection{Calibration parameter refinement\label{sec3C}}
The important factor affecting the calibration accuracy of the event cameras is their noise characteristics. Event-based sensors exhibit exceptional sensitivity towards background activity noise stemming from temporal noise and junction leakage currents, which show higher noise than other standard cameras~\cite{falanga_davide_dynamic_2020}. Moreover, line detection also leads to errors. To eliminate the negative impact caused by imaging noise and line detection, we use the linear solver to offer an initial guess. Then, we adopt non-linear refinement, e.g., gradient-based optimization, to obtain reliable calibration parameters.

For each event cluster, the 3D lines of the model can be projected onto the event plane using the projection matrix $\mathbf{M}_{i}$ at the time ${t}_{i}$. Then, establishing correspondences between the projected lines and the detected lines. The camera parameters are refined by minimizing the distance from the endpoints of detected lines to reprojected lines:

\begin{equation}
	\underset{{}}{\mathop{\arg \min }}\,\sum\limits_{j,l}{{\frac{{{\left( {{a}_{j}}x_{d}^{l}+{{b}_{j}}y_{d}^{l}+{{c}_{j}} \right)}^{2}}}{a_{j}^{2}+b_{j}^{2}}}},
	\label{eq2-13}
\end{equation}
where $({a}_{j},{b}_{j},{c}_{j})$ is the parameter of the $j^{th}$ projected line, and $(x_{d}^{l},y_{d}^{l})$ is the $l^{th}$ distorted endpoint that lies on the corresponding detected line.

To improve robustness against noise, we envisage the following scenarios where such non-linear optimization proves useful:
\begin{itemize}
	\item For single event cluster: our method only requires capturing an event cluster of the calibration object from a single viewpoint within a short period of time. After solving the initial parameters of the camera linearly, non-linear optimization is essential to further refine the camera parameters and enhance calibration accuracy.
	\item For multiple event clusters: our method also supports capturing the calibration object from multiple viewpoints, similar to other calibration methods, and then utilizes global optimization to generate the optimal intrinsic and extrinsic parameters.
	\item For minimal case solutions: our method requires at least four planar lines or six non-planar lines, which can be employed inside a robust estimation framework, such as the Random Sample Consensus (RANSAC). Then, non-linear  is used to find the best camera parameters.
\end{itemize}

\begin{figure*}[ht]
	\centering
	\includegraphics[width=1\linewidth]{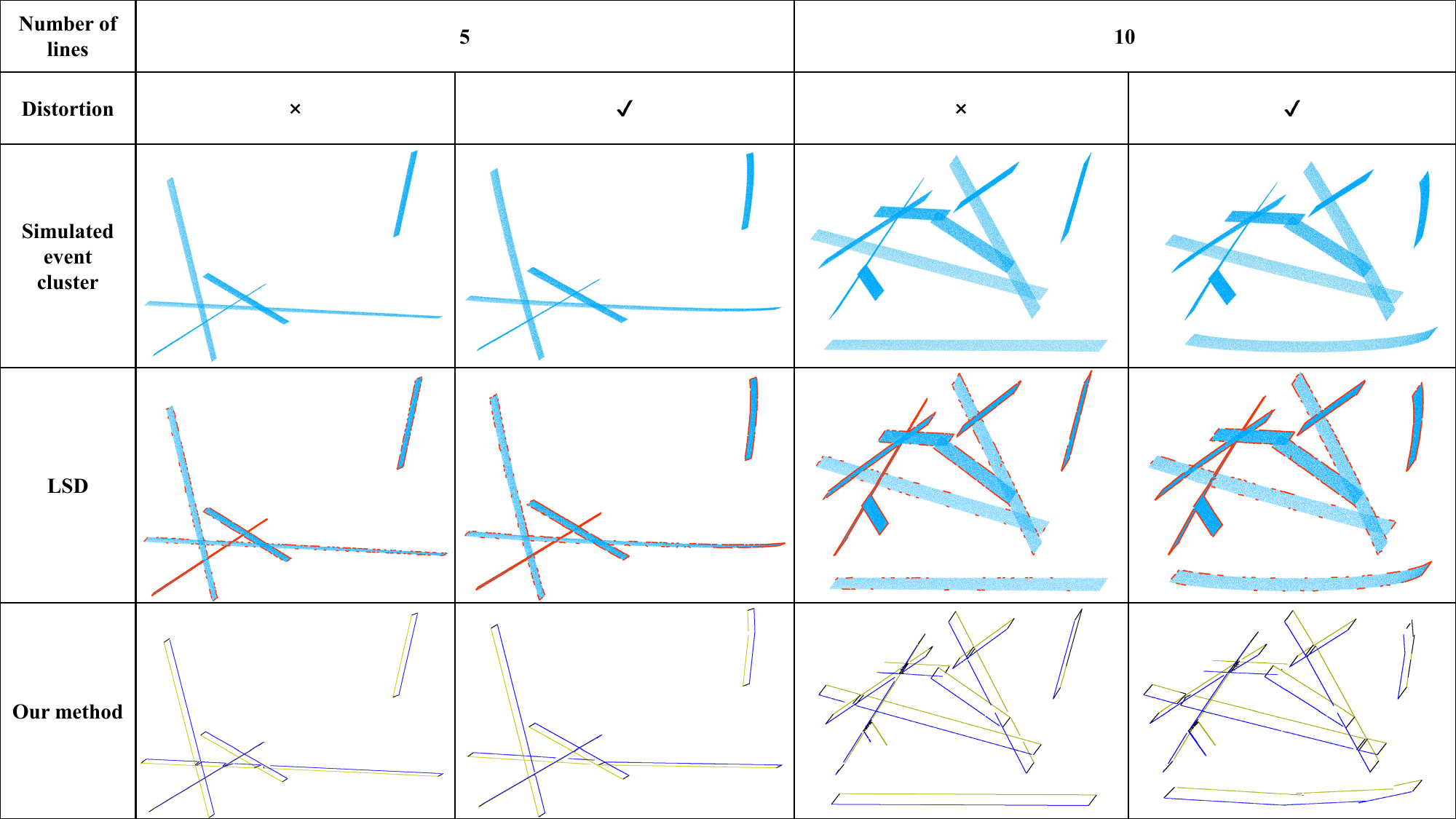}
	\caption{Line detection results of the simulated event cluster with the noise $\sigma =1$ pixels. In the third row, we present the simulated event clusters with and without distortion. The fourth row shows the results of the LSD method, where the event cluster is accumulated into a 2D image. The fifth row displays the results of our method. The extracted lines at the beginning of the cluster are shown in blue, and the ends in yellow.}
	\label{fig4}
\end{figure*}

\section{Simulated experiments\label{sec4}}
In this section, we evaluate the performance of the proposed method on simulated data, including event-based line extraction and camera calibration algorithms.
\subsection{Event-based line detection} 

To validate the effect of the event-based line detection, we randomly generate two sets of event clusters containing five and ten lines, respectively. To begin with, we generate a set of planar lines with endpoints spread in region [0,640] $\times$ [0,480] pixels on the image plane. These lines are represented by scattered points to simulate events. We then add zero-mean Gaussian noise to the events with a standard deviation of $\sigma =1$ pixels to test the robustness of the method. Subsequently, these lines move with constant angular velocity $\omega =\left[ 0,0,1 \right]\text{ rad/s}$ and linear velocity $\nu =\left[ 1,1,0 \right]$ m/s. As time progresses, these lines give rise to multiple 3D trajectories in the space-time volume to obtain the simulated event cluster. The event cluster is collected within a time interval of $0.05 $ ms, and the constant is set to ${c}_{z}=5000$. Moreover, to assess the impact of lens distortion on the method, we add a distortion coefficient  $k=\left[-0.1,-0.1\right]$ to the event cluster using the brown model. The clusters of events with and without distortion are generated and presented in Fig. 3.

We adopt the classical LSD method~\cite{von2012lsd} for comparison. Events are accumulated into a 2D image plane for line detection using the LSD method, which easily leads to confusing results, because one line triggers multiple event lines at different times. Under the interference of Gaussian noise, the method exhibits numerous missed detections. In contrast, our method detects two sets of lines, which correspond to the initial and end times of the event cluster, respectively. The line detection results of the simulated event cluster are presented at the bottom of Fig. \ref{fig4}. Although noise makes line detection difficult, our method is able to extract lines accurately. Especially in the case of lens distortion, our approach is still capable of detecting event lines, performing better than the LSD method. This is because our method performs event segmentation and reprojects them onto a 2D plane. After the projection, the edges of the curved plane remain straight, allowing for 2D line detection. Our method effectively eliminates the negative impact of distortion. The experimental results validate the feasibility of our approach, demonstrating its capability to effectively eliminate the influences caused by distortion.

\subsection{Event camera calibration} 

\begin{figure*}[t]
	\centering
	\subfloat[Median error with respect to varying image noise]{
		\begin{minipage}[t]{0.9\linewidth}
			\centering
			\includegraphics[width=1\linewidth]{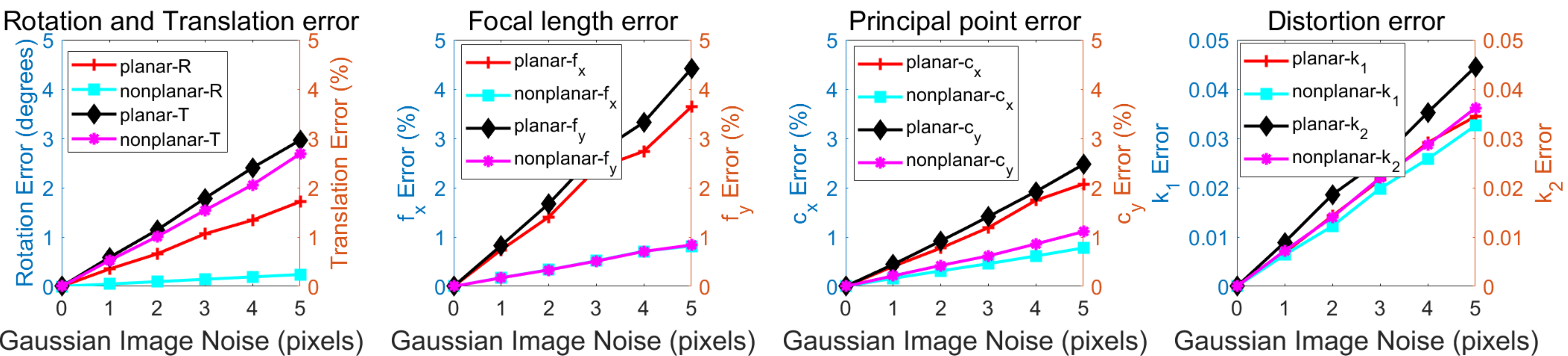}
			\label{fig3.1}
		\end{minipage}%
	}
	\\
	\subfloat[Median error with respect to varying focal length]{
		\begin{minipage}[t]{0.9\linewidth}
			\centering
			\includegraphics[width=1\linewidth]{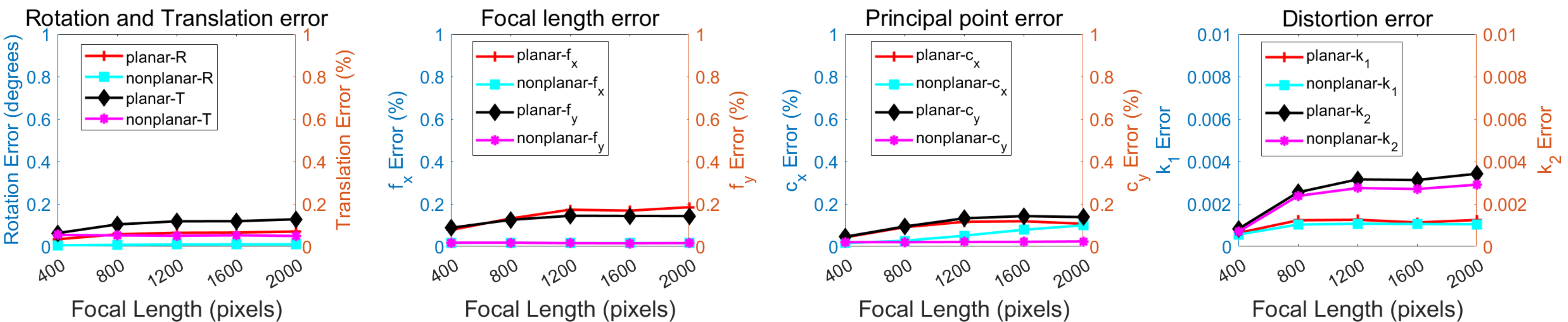}
			\label{fig3.2}
		\end{minipage}%
	}
	\\
	\subfloat[Median error with respect to varying number of lines]{
		\begin{minipage}[t]{0.9\linewidth}
			\centering
			\includegraphics[width=1\linewidth]{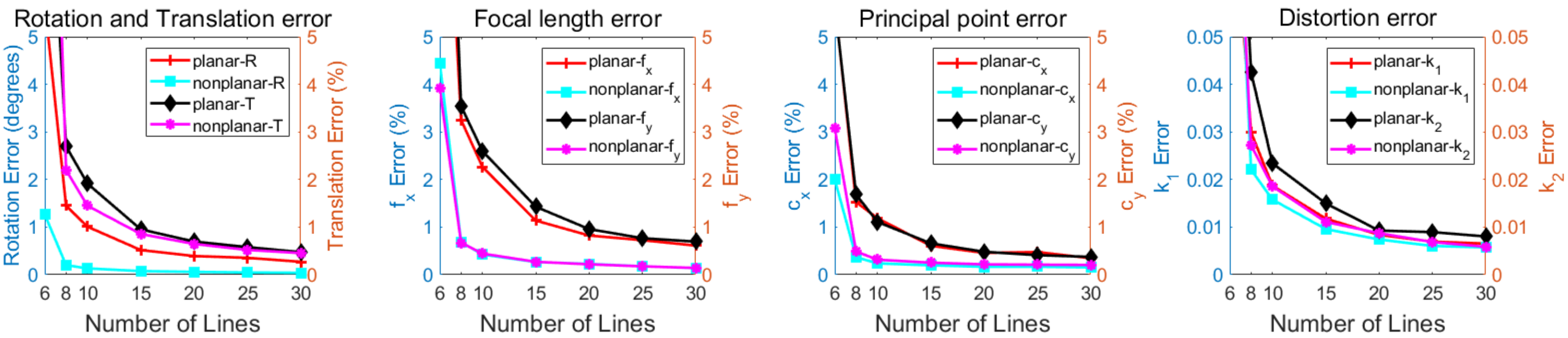}
			\label{fig3.3}
		\end{minipage}%
	}
	\\
	\subfloat[Median error with respect to varying distortion coefficients]{
		\begin{minipage}[t]{0.9\linewidth}
			\centering
			\includegraphics[width=1\linewidth]{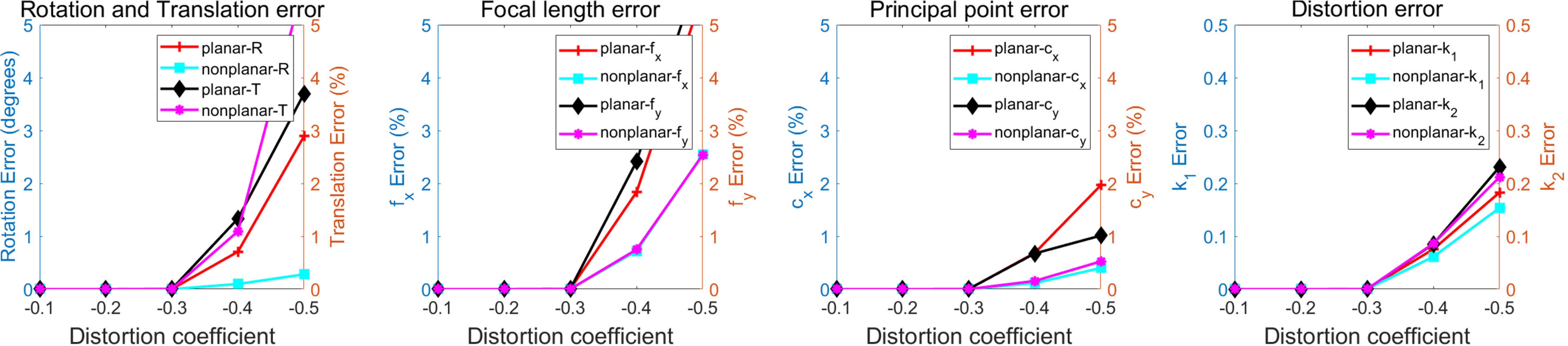}
			\label{fig3.4}
		\end{minipage}%
	}

	
	\caption{Simulation experiment results. The red and blue polylines correspond to the left axis, the black and purple polylines correspond to the right axis. (a) Median error with respect to varying image noise $\sigma$ from 0 to 5 pixels with fixed number of lines $n=25$, the focal length ${{f}_{x}}={{f}_{y}}=400$, and the principal point at the image center. (b) Median error with respect to varying focal length $f$ from 400 to 2000 pixels with fixed number of lines $n=25$, noise $\sigma=1$ pixel, and the principal point at the image center. (c) Median error with respect to varying number of lines $n$ from 6 to 30 with fixed image noise $\sigma=1$ pixel, the focal length ${{f}_{x}}={{f}_{y}}=400$, and the principal point at the image center. (d) Median error with respect to varying distortion coefficients $k_1$, $k_2$ from -0.1 to -0.5 with a step size of 0.1, with fixed image noise $\sigma=1$ pixel, the focal length ${{f}_{x}}={{f}_{y}}=400$, the number of lines $n=25$, and the principal point at the image center.}
\label{fig3}
\end{figure*}

\begin{table*}[htbp]
	\centering
	\caption{Ablation studies of our method.}
	\setlength{\tabcolsep}{4.5pt}
	\begin{tabular}{cccccccccc}
		\toprule
		Method  & Step  & $Err_\mathbf{R}$(°)     & $Err_\mathbf{T}$(\%)  & $Err_{f_x}$(\%)    & $Err_{f_y}$(\%)    & $Err_{c_x}$(\%)    & $Err_{c_y}$(\%)    & $Err_{k_1}$(${{10}^{-3}}$)   & $Err_{k_2}$(${{10}^{-3}}$) \\
		\midrule
		\multirow{2}[2]{*}{Our-planar} & Initial & 2.47  & 5.09  & 4.25  & 5.40  &$\setminus$     & $\setminus$    & 6.86   & 8.91 \\
		& Optimization & 0.35  & 0.58  & 0.74  & 0.89  & 0.40  & 0.44  & 5.98 & 8.12 \\
		\midrule
		\multirow{2}[2]{*}{Our-nonplanar} & Initial & 1.49  & 4.70  & 5.01   & 4.54   & 2.25  & 2.78  & 6.41   & 7.22 \\
		& Optimization & 0.05  & 0.52  & 0.18  & 0.17  & 0.16  & 0.21  & 5.98 & 6.67\\
		\bottomrule
	\end{tabular}%
	\label{table0}%
\end{table*}%

Given a virtual event camera with image resolution $640\times 480$ pixels, focal length ${{f}_{x}}={{f}_{y}}=400$ pixels, and the principal point at the image center. Considering the real situation, distortion coefficients are set to $k=\left[-0.1,-0.1\right]$, which ignores the high-order coefficients. The event lines are randomly generated and then projected into the world coordinate system to obtain planar and non-planar 3D lines. The rotation and translation of the event camera are randomly determined on each trial, and we perform 1000 independent trials for each test. 
%
We calculate the error of the focal length and principal point relative to the true value in percent. The distortion error is defined by the absolute error, and the error metric of the rotation matrix and translation vector are calculated as the same as in~\cite{wang_camera_2019,xu_pose_2017}:
\begin{equation}
Er{{r}_{\mathbf{R}}}=\max \left\{ {\arccos }\left( r_\text{g,truth}^{T}{{r}_\text{g}} \right)\times {180}/{\pi }\; \right\}\left( g=1,2,3 \right),
\label{eq2-14}
\end{equation}

\begin{equation}
Er{{r}_{\mathbf{T}}}={\left\| \mathbf{T}-{{\mathbf{T}}_{\text{truth}}} \right\|}/{{{\mathbf{T}}_{\text{truth}}}},
\label{eq2-15}
\end{equation}
where ${{r}_\text{g}}$ and ${{r}_\text{g,truth}}$ are the $g_{th}$ column vectord of the estimated value $\mathbf{R}$ and the true value ${{\mathbf{R}}_\text{truth}}$, respectively. According to~\cite{xu_pose_2017}, the results are considered as correct when $Er{{r}_{\mathbf{R}}}<5\text{ }degrees$ and $Er{{r}_{\mathbf{T}}}<5\text{ }percent$.

The methods employed for the camera calibration using planar and non-planar lines are abbreviated as "Our-planar" and "Our-nonplanar", respectively. In order to provide a more intuitive comparison, we utilize "planar/nonplanar-X" as a notation to represent the errors of Our-planar/nonplanar methods,  with "X" signifying the camera parameters.

\textbf{Accuracy with respect to varying image noise:}
The first experiment is designed to test the effect of noise on the accuracy of our method. We fix the number of lines $n=25$, the focal length ${{f}_{x}}={{f}_{y}}=400$, the principal point at the image center (center aligned lens and camera). Then, we add zero-mean Gaussian noise to the endpoints of the 2D lines with a standard deviation $\sigma$ varying from 0 to 5 pixels. The median error of the parameter estimation accuracy is shown in Fig. 4(a), which almost grows linearly as the noise increases. Overall, Our-nonplanar method has higher accuracy compared to Our-planar method in the noise case, especially in the focal length error of Fig. 4(a).

\textbf{Accuracy with respect to varying focal length:}
In the second experiment, the focal length $({{f}_{x}},{{f}_{y}})$ varies from 400 to 2000 pixels. The number of lines is set to $n=25$, the standard deviation of Gaussian noise is fixed to $\sigma=1$ pixel, and the principal point is at the image center. Fig. 4(b) shows that most parameters are relatively stable in both planar and non-planar cases, and further indicates that our methods achieve satisfactory performance with varying focal lengths.

\begin{figure*}[h]
	\centering
	\includegraphics[width=1\linewidth]{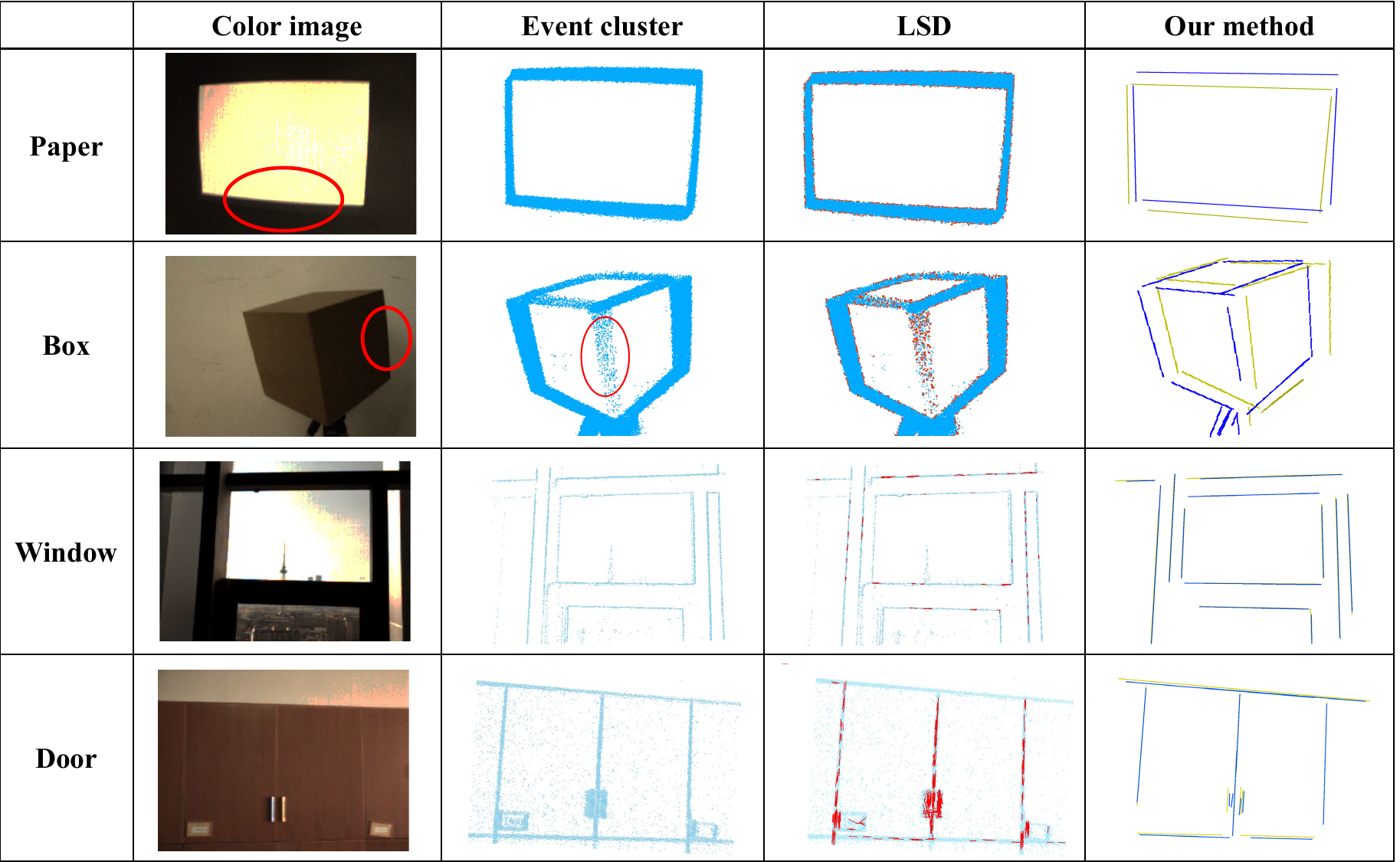}
	\caption{Line extraction results of the real data. The second column displays the color image captured by DAVIS34, where lens distortion is observed at the edges, as marked in the red circle area. The third column displays the event clusters, which have already been accumulated onto the event plane. The fourth column displays the line detection results using the LSD method, marked in red. The fifth column shows the line detection results obtained by our method. The extracted lines at the beginning of the cluster are shown in blue, and the ends in yellow.}
	\label{fig5}
\end{figure*}

\textbf{Accuracy with respect to varying number of lines:}
In the next experiment, we investigate the accuracy with respect to the varying number of lines under a fixed image noise $\sigma =1$ pixel, the focal length ${{f}_{x}}={{f}_{y}}=400$, and the principal point at the image center. The number of lines varies from 6 to 30, which is suitable for both planar and non-planar cases. As indicated in Fig. 4(c), when $n\ge8 $, Our-nonplanar method achieves reliable results. With the increasing number of lines, the parameter errors of Our-planar and Our-nonplanar methods gradually stabilize.

\textbf{Accuracy with respect to varying distortion coefficients:}
In this experiment, we investigate the impact of variations in distortion coefficients on the accuracy of camera calibration. The distortion coefficients $k_1$ and $k_2$ varied from -0.1 to -0.5 with a step size of 0.1, with fixed image noise $\sigma=1$ pixel, the focal length ${{f}_{x}}={{f}_{y}}=400$, the number of lines $n=25$, and the principal point at the image center. The experimental results are illustrated in Fig. 4(d). When the coefficient of distortion is minimal, the overall error is exceedingly insignificant; however, as the distortion progressively magnifies, the error concomitantly amplifies, especially for the rotation, translation and focal length error. The variation in distortion coefficients directly affects the accuracy of the initial geometric solution for calibration. Consequently, we utilize nonlinear optimization to refine these camera parameters, which can effectively reduce the interference caused by distortion.


\textbf{Ablation Study:} 
To evaluate each component of the proposed method, we divide it into two parts, namely, initial calibration and optimization, and separately scrutinize the calibration precision of these two segments. The simulation parameters remain consistent, with image noise $\sigma=1$ pixel, the focal length ${{f}_{x}}={{f}_{y}}=400$, number of lines $n=25$, and the principal point at the image center. The experimental results of the ablation study are presented in Table~\ref{table0}, using the median error of the calibration parameters. Due to the influence of noise and distortion, the initial calibration error for geometric calculation is relatively high. However, after nonlinear optimization, a significant improvement in accuracy can be observed. This means that it is effective to employ our optimization method to refine camera calibration parameters.

\section{Real-world experiments}\label{sec5}
In order to demonstrate the robustness and generality of our methods, we conduct a series of experiments, including line detection, monocular camera calibration, and stereo camera calibration. 
For the purpose of comparison, we test two kinds of event cameras: DAVIS346 (resolution 346 $ \times $ 260 pixels) and Prophesee EVK4 (resolution $ 1280\times 720$ pixels). Moreover, we adopt an A4-size paper, a box, windows and doors as the calibration reference objects, all possessing planar or non-planar line features. The reason for selecting these objects is that they are quite common in man-made environments, which demonstrates the operational simplicity and broad applicability of our methods.

\subsection{Event-based line detection} 
Firstly, line extraction is performed on the four calibration reference objects. We use the DAVIS346 to obtain the event cluster and then detect lines using our method. Fig. \ref{fig5} shows the local details of the line detection result. Events contain so much noise that only a few lines are detected by the LSD method. In contrast, our method has successfully detected the edge lines of the objects at the beginning and the end of each event cluster, which exhibits greater robustness. During the actual imaging process, events are also affected by lens distortion. Fig.~\ref{fig5} vividly illustrates event lines that are curved due to distortion, as marked in the red circle area. Despite this, we are still able to detect all distorted lines, demonstrating the feasibility of our method in practical scenarios.

\begin{table}[tbp]
	\centering
	\newcommand{\tabincell}[2]{\begin{tabular}{@{}#1@{}}#2\end{tabular}}
	\caption{Event camera (DAVIS346) calibration results \uppercase\expandafter{\romannumeral1}.}
	\setlength{\tabcolsep}{4pt}
	\begin{tabular}{ccccc}
		\toprule
		& \multicolumn{1}{c}{Method} & \multicolumn{1}{c}{{Toolbox~\cite{toolboxdavis}}} & \multicolumn{1}{c}{{Our-planar}} & \multicolumn{1}{c}{{Our-nonplanar}} \\
		\midrule
		& \multicolumn{1}{c}{{{\tabincell{c}{Calibration\\object}}}} & \multicolumn{1}{c}{{Checkerboard}} & \multicolumn{1}{c}{{Paper}} & \multicolumn{1}{c}{{Box}} \\
		\midrule
		& ${f}_{x}$   & 320.11 & 315.24 & 326.52  \\
		& ${f}_{y}$   & 319.01 & 311.16 & 324.63  \\
		& ${c}_{x}$   & 161.38 & 173.00 & 164.95  \\
		& ${c}_{y}$   & 133.14 & 130.00 & 138.15  \\
		& ${k}_{1}$   & -0.27  & -0.21  & -0.19   \\
		\bottomrule
	\end{tabular}%
	\label{table1}%
\end{table}%

\begin{table}[tbp]
	\centering
	\newcommand{\tabincell}[2]{\begin{tabular}{@{}#1@{}}#2\end{tabular}}
	\caption{Event camera (DAVIS346) calibration results \uppercase\expandafter{\romannumeral2}.}
	\setlength{\tabcolsep}{5.7pt}
	\begin{tabular}{ccccc}
		\toprule
		& \multicolumn{1}{c}{Method} & \multicolumn{1}{c}{{E2Calib~\cite{muglikar_how_2021}}} & \multicolumn{1}{c}{{Our-planar}} & \multicolumn{1}{c}{{Our-planar}} \\
		\midrule
		& \multicolumn{1}{c}{{{\tabincell{c}{Calibration\\object}}}} & \multicolumn{1}{c}{{Checkerboard}} & \multicolumn{1}{c}{{Window}} & \multicolumn{1}{c}{{Door}} \\
		\midrule
		& ${f}_{x}$   & 471.95 & 479,12 & 468.58  \\
		& ${f}_{y}$   & 471.93 & 467.03 & 475.82  \\
		& ${c}_{x}$   & 172.99 & 173.00 & 173.00  \\
		& ${c}_{y}$    & 124.16 & 130.00 & 130.00\\
		& ${k}_{1}$   & -0.10  & -0.12  & -0.03 \\
		\bottomrule
	\end{tabular}%
	\label{table2}%
\end{table}%

\begin{figure*}[tbp]
\centering
\subfloat[Paper]{
	\begin{minipage}[t]{0.24\linewidth}
		\centering
		\includegraphics[width=1\linewidth]{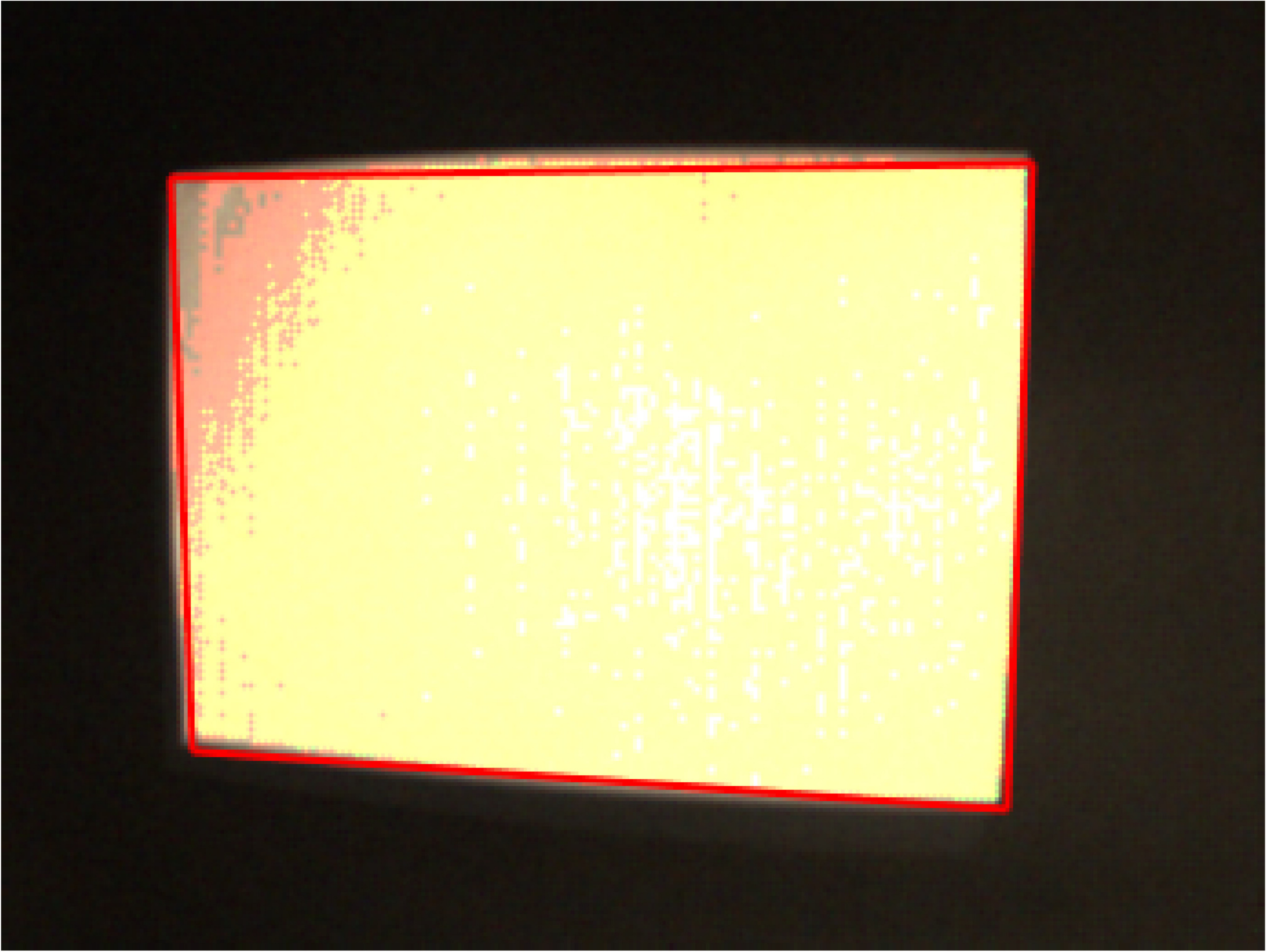}
		\label{fig6.1}
	\end{minipage}%
}
\subfloat[Box]{
	\begin{minipage}[t]{0.24\linewidth}
		\centering
		\includegraphics[width=1\linewidth]{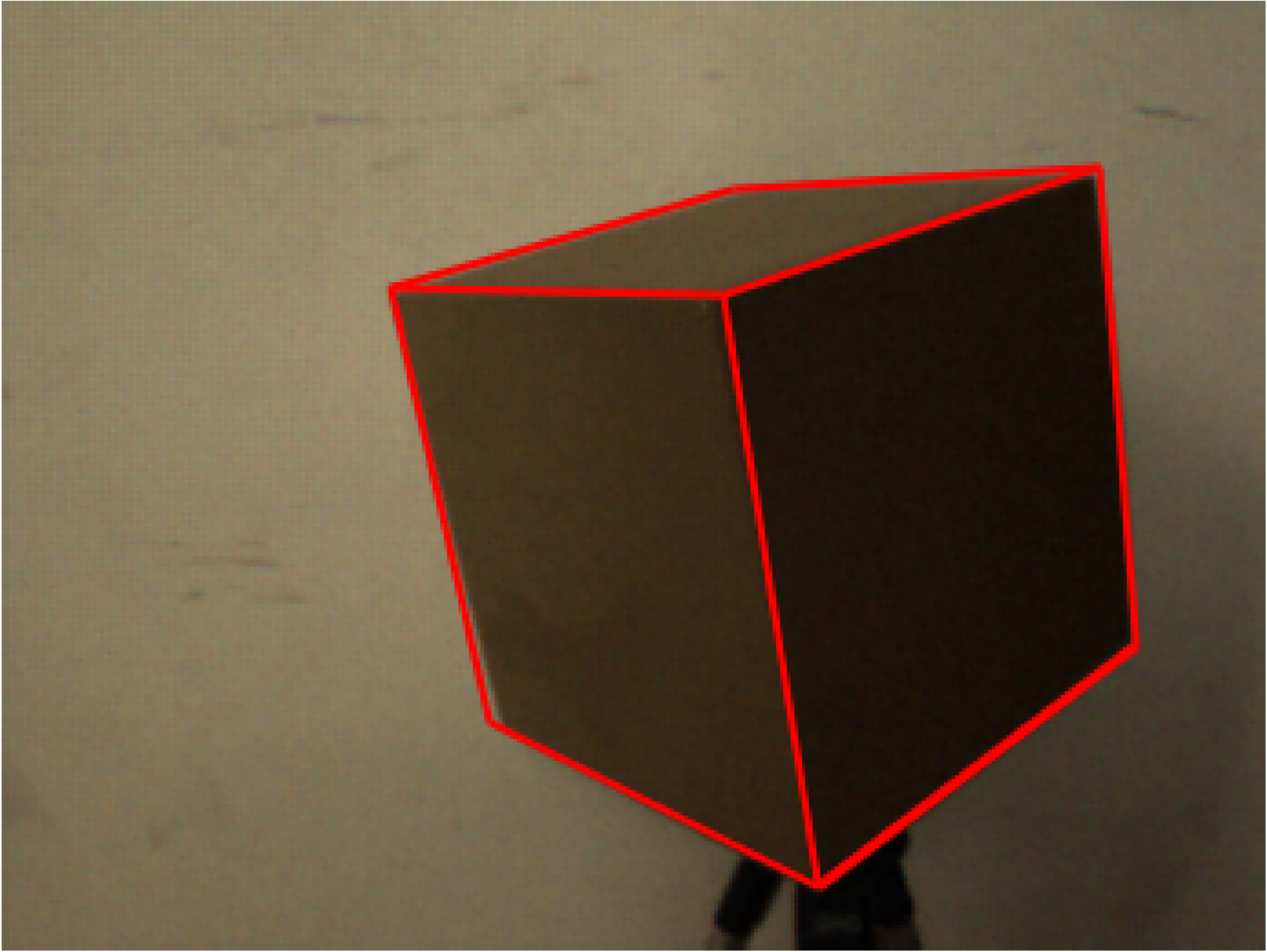}
		\label{fig6.2}
	\end{minipage}%
}
\subfloat[Window]{
	\begin{minipage}[t]{0.24\linewidth}
		\centering
		\includegraphics[width=1\linewidth]{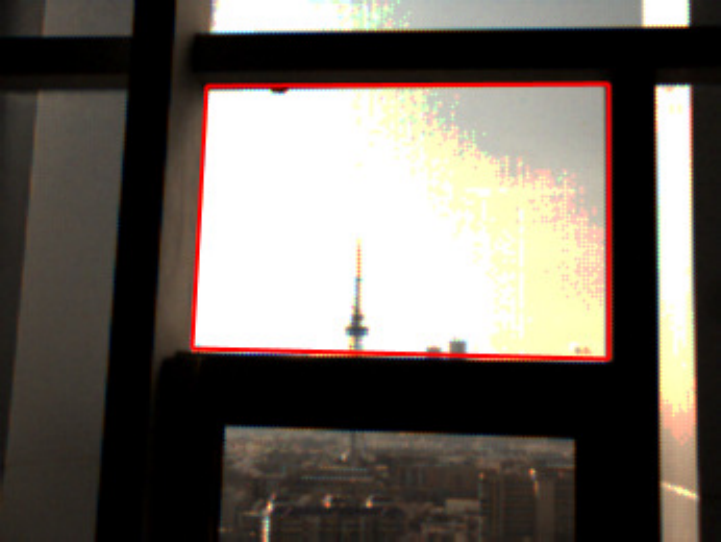}
		\label{fig6.3}
	\end{minipage}%
}
\subfloat[Door]{
	\begin{minipage}[t]{0.24\linewidth}
		\centering
		\includegraphics[width=1\linewidth]{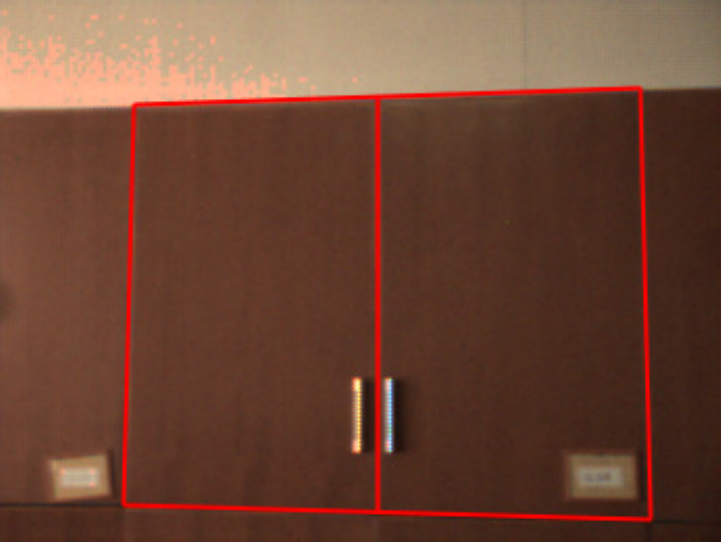}
		\label{fig6.4}
	\end{minipage}%
}
\caption{Visualization of reprojection results of DAVIS346. 3D lines are reprojected onto the images (indicated by the red lines).}
\label{fig6}
\end{figure*}

\begin{table*}[t]
\centering
\newcommand{\tabincell}[2]{\begin{tabular}{@{}#1@{}}#2\end{tabular}}
\caption{Stereo event camera (Prophesee EVK4) calibration results \uppercase\expandafter{\romannumeral1}.}
\setlength{\tabcolsep}{8pt}
\begin{tabular}{ccccc}
	\toprule
	& {Method} & {Toolbox~\cite{bouguet2004camera}} & {Our-planar} & {Our-nonplanar} \\
	\cmidrule{2-5}
	&{\tabincell{c}{Calibration object}}&{Checkerboard} & {Paper} & {Box} \\
	\midrule
	\multirow{3}[2]{*}{{\tabincell{c}{Intrinsic parameters\\(left camera)}}} & $(f_x,f_y)$  &    (1663.02,1662.42)   &  (1647.44,1680.77)     &  (1667.69,1682.17)\\
	& $(c_x,c_y)$  &  (635.28,357.22)     &   (640.00,360.00)    &  (639.64,362.83)\\
	& $k_1$  &   -0.09    &  -0.05    & -0.01 \\
	\midrule
	\multirow{3}[2]{*}{{\tabincell{c}{Intrinsic parameters\\(right camera)}}}  & $(f_x,f_y)$  &  (1643.29,1642.78)     &   (1652.90,1620.85)    &  (1653.02,1659.69)\\
	& $(c_x,c_y)$  &   (619.81,372.05)    &   (640.00,360.00)    &  (617.73,358.42)\\
	& $k_1$  &    -0.07   &   -0.02    &  -0.02\\
	\midrule
	\multirow{2}[2]{*}{{\tabincell{c}{Extrinsic parameters}}} & $\mathbf{R_{r2l}}\left( ^\circ  \right)$     & [-1.86,-2.21,2.33] &  [-1.87,-2.21,3.40] &  [-1.83,-2.07,1.37] \\
	& ${{\mathbf{T}}_{\mathbf{r2l}}}\left( cm \right)$  &    [16.18,1.86,3.74]   &   [14.42,1.29,1.46]    &  [15.14,2.71,4.39] \\
	\bottomrule
\end{tabular}%
\label{tabel3}%
\end{table*}%

\begin{table*}[t]
\centering
\newcommand{\tabincell}[2]{\begin{tabular}{@{}#1@{}}#2\end{tabular}}
\caption{Stereo event camera (Prophesee EVK4) calibration results \uppercase\expandafter{\romannumeral2}.}
\setlength{\tabcolsep}{8pt}
\begin{tabular}{ccccc}
	\toprule
	& {Method} & {Flashing screen~\cite{reinbacher2017real}} & {Our-planar} & {Our-planar} \\
	\cmidrule{2-5}
	&{\tabincell{c}{Calibration object}} &{Checkerboard} & {Window} & {Door} \\
	\midrule
	\multirow{3}[2]{*}{{\tabincell{c}{Intrinsic parameters\\(left camera)}}} & $(f_x,f_y)$  &    (1693.53,1695.33)   &  (1708.83,1718.30)     &  (1706.43,1709.92)\\
	& $(c_x,c_y)$  &  (636.92,350.57)     &   (640.00,360.00)    &  (640.00,360.00)\\
	& $k_1$  &   -0.09    &  -0.13    & -0.04 \\
	\midrule
	\multirow{3}[2]{*}{{\tabincell{c}{Intrinsic parameters\\(right camera)}}}  & $(f_x,f_y)$  &   (1709.17,1714.29)      &   (1688.98,1687.60)    &  (1694.40,1695.00)\\
	& $(c_x,c_y)$  &   (646.61,357.28)    &   (640.00,360.00)    &  (640.00,360.00)\\
	& $k_1$  &    -0.03   &   -0.05    &  -0.04\\
	\midrule
	\multirow{2}[2]{*}{{\tabincell{c}{Extrinsic parameters}}} & $\mathbf{R_{r2l}}\left( ^\circ  \right)$     & [1.86,-13.45,-6.73] &  [1.77,-13.66,-7.12] & [1.69,-12.87,-6.00] \\
	& ${{\mathbf{T}}_{\mathbf{r2l}}}\left( cm \right)$    &    [18.54,-1.38,3.27]   &   [19.49,-1.45,3.73]    &  [17.37,-1.51,4.78] \\
	\bottomrule
\end{tabular}%
\label{tabel4}%
\end{table*}%

\subsection{Monocular event camera calibration}

We conduct monocular camera calibration experiments. DAVIS346 is able to capture regular RGB images of a checkerboard at a frame rate of 30 Hz, which allows us to compare our method with classical image-based calibration methods. To ensure calibration accuracy, we capture thirty RGB images of a checkerboard from different angles using DAVIS346 and perform calibration using the toolbox from DV~\cite{toolboxdavis}. In addition, we compare our method with E2calib~\cite{muglikar_how_2021}, which converts events into grayscale images and utilizes traditional calibration algorithms for calibration. Unlike conventional algorithms, our method directly leverages event streams for calibration. Our-planar method is validated using paper, window and door, and Our-nonplanar method is tested by a box, both of which utilize commonly-encountered objects in man-made environments. 


In the monocular camera calibration experiments, our primary comparison revolved around the camera's intrinsic parameters, encompassing focal length, principal point, and distortion coefficients. We conduct two sets of calibration experiments on DAVIS346, using different lenses, and the results are recorded in Table~\ref{table1} and Table~\ref{table2}, respectively. As can be easily observed, our method yields intrinsic parameters with a precision comparable to that achieved by the classical method, while only a minimum of four lines are required. Our method only needs an event cluster occurring within a short time interval, whereas the toolbox~\cite{toolboxdavis} requires capturing multiple images of checkerboard from different angles. To demonstrate the accuracy of the extrinsic parameters, we reproject 3D lines of the calibration objects on the image (provided by DAVIS346) at the time $t_{i}$, utilizing the camera intrinsic and extrinsic parameters. As demonstrated in Fig.~\ref{fig6}, the reprojection results are visually compelling, as the reprojected lines fit well with the object edges.

\subsection{Stereo event camera calibration}

In this section, we evaluate the performance of our calibration method on stereo event cameras. The calibration objects used in this experiment are the same as the ones used in the previous experiment. We utilize two stationary Prophesee EVK4 event cameras to capture the chessboard for the acquisition of event streams.
Unlike the DAVIS346, the EVK4 cameras do not have the capability to directly obtain RGB images. Consequently, we plan to convert event streams into multiple 2D images and perform calibration using the toolbox from Bouguet~\cite{bouguet2004camera}. Specifically, events occurring within a short time interval are aggregated onto a grayscale image. This method captures the chessboard from various angles and obtains multiple grayscale images for calibration, resulting in a high level of accuracy. Furthermore, inspired by~\cite{reinbacher2017real}, we employ a screen displaying a chessboard pattern image that alternates rapidly with a blank frame. This allowed us to generate grayscale images that serve as inputs for conventional camera calibration pipelines. These two methods both collected thirty images for calibration, and the camera parameters they obtained are regarded as the baseline for comparing the accuracy of our method.

Similar to the DAVIS346 calibration experiment, we conduct two sets of calibration experiments on EKV4 cameras using different lenses.  Table~\ref{tabel3} and Table~\ref{tabel4} report the calibration results of the stereo event camera, including the intrinsic and extrinsic parameters. For a more intuitive presentation of the results, we converted the rotation matrix $\mathbf{R_{r2l}}$ to rotation angles using the Rodrigues transformation. Generally, our methods yield reliable calibration results by utilizing the events from a single perspective within a short time interval. Compared to traditional methods that require multiple images, our methods are more convenient and efficient. Our-nonplanar method provides better accuracy of camera parameters in comparison to Our-planar method. Because nonplanar lines can provide 3D information, leading to a more stable computation of the camera projection matrix and resulting in a more accurate estimation of camera intrinsic and extrinsic parameters.

In practical applications, we recommend flexibly selecting these two methods based on different application scenarios. For the sake of calibration accuracy, the preferred method is Our-nonplanar method, which requires at least six nonplanar lines, particularly in the presence of high levels of event noise. If applied in a constrained environment with only a few planar lines available (four or more), Our-planar method can also be used for camera calibration. 

\section{Conclusion\label{sec6}}
We propose an event camera calibration method using geometric lines. First, we detect lines directly from event clusters, overcoming the interference caused by distortion that results in straight lines appearing curved. Second, we estimate initial camera parameters, relying on the line-based perspective projection model as well as the event motion model. Furthermore, enabling batch optimization to improve performance. Simulated and real-world experiments demonstrate our framework achieves satisfying calibration results for both monocular and stereo event cameras. Our method can be easily extended to calibrate other electro-optical sensors, e.g., ToF cameras and RGB-D cameras. Future work involves exploring the applicability of our method in other computer vision tasks, including but not limited to pose estimation, 3D reconstruction, and virtual reality, all intimately related to the problem of event-based camera calibration.

\section{Acknowledgments}
This work was supported in part by the National Natural Science Foundation of China under Grant 12372189, and the Hunan Provincial Natural Science Foundation for Excellent Young Scholars under Grant 2023JJ20045.









\printcredits

\bibliographystyle{cas-model2-names}

\bibliography{refe}



\end{document}